\documentclass[preprint,12pt]{elsarticle}

\usepackage{hyperref}
\usepackage{epsfig}
\usepackage{graphicx}
\usepackage{algorithm}
\usepackage{algorithmic}
\usepackage{amssymb}
\usepackage{float}
\usepackage{enumitem}
\usepackage{subfigure}
\usepackage{booktabs}
\usepackage{multirow}

\journal{Pattern Recognition}

\begin{document}

\begin{frontmatter}



\title{A Distributed Deep Representation Learning Model \\for Big Image Data Classification}


\author[mymainaddress]{Le Dong\corref{mycorrespondingauthor}}
\cortext[mycorrespondingauthor]{Corresponding author. Tel.:+86 13981763623; Fax: +86-28-61831655.}
\ead{ledong@uestc.edu.cn}

\author[mymainaddress]{Na Lv}

\author[mysecondaryaddress]{Qianni Zhang}

\author[mymainaddress]{Shanshan Xie}

\author[mymainaddress]{Ling He}

\author[mymainaddress]{Mengdie Mao}

\address[mymainaddress]{School of Computer Science and Engineering, University of Electronic Science and
Technology of China (UESTC),2006 Xiyuan Avenue, Gaoxin West Zone, Chengdu,
Sichuan, 611731, China}
\address[mysecondaryaddress]{School of Electronic Engineering and Computer Science, Queen Mary, University of
London}

\begin{abstract}
This paper describes an effective and efficient image classification framework nominated distributed deep representation learning model (DDRL).  The aim is to strike the balance between the computational intensive deep learning approaches (tuned parameters) which are intended for distributed computing, and the approaches that focused on the designed parameters but often limited by sequential computing and cannot scale up. In the evaluation of our approach, it is shown that DDRL is able to achieve state-of-art classification accuracy efficiently on both medium and large datasets. The result implies that our approach is more efficient than the conventional deep learning approaches, and can be applied to big data that is too complex for parameter designing focused approaches. More specifically, DDRL contains two main components, \emph{i.e.}, feature extraction and selection. A hierarchical distributed deep representation learning algorithm is designed to extract image statistics and a nonlinear mapping algorithm is used to map the inherent statistics into abstract features. Both algorithms are carefully designed to avoid millions of parameters tuning. This leads to a more compact solution for image classification of big data. We note that the proposed approach is designed to be friendly with parallel computing. It is generic and easy to be deployed to different distributed computing resources. In the experiments, the large-scale image datasets are classified with a DDRM implementation on Hadoop MapReduce, which shows high scalability and resilience.
\end{abstract}


\end{frontmatter}

\section{Introduction}
{R}{ecent} years have witnessed great success and the
 development of deep learning \cite{Hinton1} applied to multiple
 levels of representation and abstraction that help make sense
 of image data to accomplish higher-level tasks such as image retrieval \cite{Liang37, Penatti4},
classification \cite{Samat2, Dong38, Dong39}, detection \cite{Li3, Pedersoli40}, \emph{etc.} Elegant
 deep representation obtained through greedily learning
 successive layers of features will contribute to make
 subsequent tasks more achievable. Provided the scarce
 labeled data, current deep learning methods such as CNN (Convolutional Neural Networks) \cite{Krizhevsky5, Jarrett6},
 Sparse coding \cite{Yang7}, Sparse auto-encoder \cite{Goodfellow8,Poultney11} and RBMs (Restricted Boltzmann Machines) \cite{Krizhevsky10}
 typically employed an unsupervised learning algorithm to train a
 model of the unlabeled data and then used the gained deep
 representation to extract interesting features. These
 aforementioned deep learning models generally have huge amounts
 of hyper-parameters to be tuned, which impose sharp requirements
 for storage and computational expense. More recently, researchers
 found that it is possible to achieve state-of-the-art performance
 by focusing effort on the design parameters (\emph{e.g.}, the receptive
 field size, the number of hidden nodes, the step-size between
 extracted features, \emph{etc.}) with simple learning
 algorithms and a single layer of features \cite{Coates9}. However, the
 superiority demonstrated in \cite{Coates9} is based on
the relatively small benchmarkdatasets like NORB \cite{Deng12} and CIFAR-100 \cite{Krizhevsky10}.
When applied to big image datasets such
as ImageNet \cite{Dean13}, the classification accuracy of the single layer feature approach
may suffer from the information loss during the feature extraction. Indeed, big image
data comes along accompanied by the widespread real applications
in various areas, such as engineering, industrial manufacture,
military and medicine, \emph{etc.}, which directly motivate us to construct a robust and reliable model for big image data
classification with the joining efforts of feature design, deep learning and
distributed computing resources.

Inspired by the previous state-of-the-art approaches \cite{Coates9, Wu18, Coates36}, we
utilize hierarchical distributed deep representation learning algorithm, based on K-means, to serve as the unsupervised
feature learning module. The proposed approach avoids the selection
of multiple hyper-parameters, such as learning rates, momentum,
sparsity penalties, weight decay which must
be chosen through cross-validation and result in
substantially increased runtime. K-means has enjoyed wide adoption
in computer vision for building codebooks of visual words
used to define higher-level image features, but it has been
less widely used in deep learning. In our design,
K-means is used to construct a dictionary $D\in R^{n\times k}$
of $k$ centroids in each layer to gain the feature mapping function
so that an input data vector $x^{(i)} \in R^n, i = 1, \cdots, m$
can be mapped to a new feature representation that minimizes
the error in reconstruction. The proposed approach is computational simpler as
not require any hyper-parameter to be tuned other than
obtaining the dictionary $D$. We note that the complexity of the dictionary grows linearly
with the number of the layers, which imposes non-trivial computations cannot be
handled by a single machine. To mitigate this problem, we utilize
the distributed computing resources to provide competent
computing capability and storage. Here, the prevalent
MapReduce \cite{Dean13}, aimed at big data parallel processing, is
chosen to serve as the implementation platform. Based on this platform, our
proposed Distributed Deep Representation Learning
Model (DDRL) is reliably trained. Note that DDRL model is not
restricted to be run on MapReduce. It is generic and
can be deployed to any other distributed platform.

In general, the significant contribution of this paper can be summarized as three aspects:

1) The proposed DDRL model is a hierarchical structure designed to abstract the layer-wise image feature information. Each hierarchy is based on hierarchical distributed deep representation learning algorithm to learn the inherent statistics of the image data. The DDRL model abandons millions of parameters estimation, which releases the complexity of the traditional model learning methods;

2) To further counter the computing challenges
brought by big image data, DDRL model is set up on the distributed resources which help to release the storage and computation efficiency issues. In addition, each layer adapts parallel processing, which further improves the scalability and fault tolerance of DDRL;

3) Owning to the excellent parallel design and simplifying burdensome models, our DDRL model learns the saliency representation to achieve big image data classification and obtains desirable performance.

The remainder of this paper is organized as follows: Section 2
provides a review of the related works and Section 3 elaborates
our proposed approach. Experimental evidences that validates
our work and a comparison with other methods are presented in Section
4. Finally, Section 5 concludes the paper.

\section{Related work}

In recent years, much attention has been concentrated on
the flourish of deep learning, which can be of
unsupervised \cite{Hinton1}, supervised \cite{Deng12}, or a hybrid form \cite{Liu17}.
Hierarchical and recursive networks \cite{Hinton1, Huang18} have demonstrated
great promise in automatically learning thousands or even millions
of features. Image classification \cite{Hinton19, Ngiam20, LuoY32, Fu33, Babu34, Simonyan35} based on deep learning
have also observed significant performance, especially in the
presence of large amount of training data. \cite{Simonyan35} extracts hand-designed low-level features, which fails to capture the inherent variability of images.

Despite the superiority of deep learning on processing
vision tasks, some potential problems with the current deep
learning frameworks still exist, just as \cite{Lee21} concluded: reduced transparency and determinativeness of the features
learned at hidden layers \cite{Zeiler22}; training difficulty due to
exploding and vanishing gradients \cite{Pascanu23, Glorot24}; lack of a thorough
mathematical understanding about the algorithmic behavior,
despite of some attempts made on the theoretical side \cite{Eigen25};
dependence on the availability of large amount of training
data \cite{Hinton19}; complexity of manual tuning during training \cite{Krizhevsky5}.
To enhance the performance of deep learning from various angles,
several techniques such as dropout \cite{Hinton19}, drop connect \cite{Wan26},
pre-training \cite{Dahl27}, and data augmentation \cite{Ciresan28}, have been proposed.
In addition, a variety of engineering tricks are employed to
fine-tune feature scale, step size, and convergence rate.

In \cite{Coates9}, k-means successfully plays an unsupervised feature
learning role to achieve good performance. It is particularly
noteworthy for its simple implementation and fast training.
Unfortunately, it suffers from the problem that a very large
number of centroids are required to generate good features,
which directly brings heavy burden on computing speed and
storage capacity. To take advantages of the simpleness of
K-means and overcome aforementioned deficiencies, we consider
employing the distributed resources to make contributions.
In this sense, MapReduce \cite{Dean13, Yu29}, as a prevalent distributed
processing framework, is a reliable platform to provide
sufficient computing resources. MapReduce is a prevalent
framework capable of efficiently processing huge data amount
in a parallel manner across numerous distributed nodes. The
excellent fault tolerance and load balance of MapReduce
benefit from its inherent working mechanism which detects
failed map or reduces tasks and reschedules these tasks to
other nodes in the cluster. Detailed operation and combination
of MapReduce with our DDRL model will be further
introduced in subsequent sections.
\begin{figure}[H]
\centering
\scalebox{1.2}[1.2]{\includegraphics[width=0.85\linewidth]{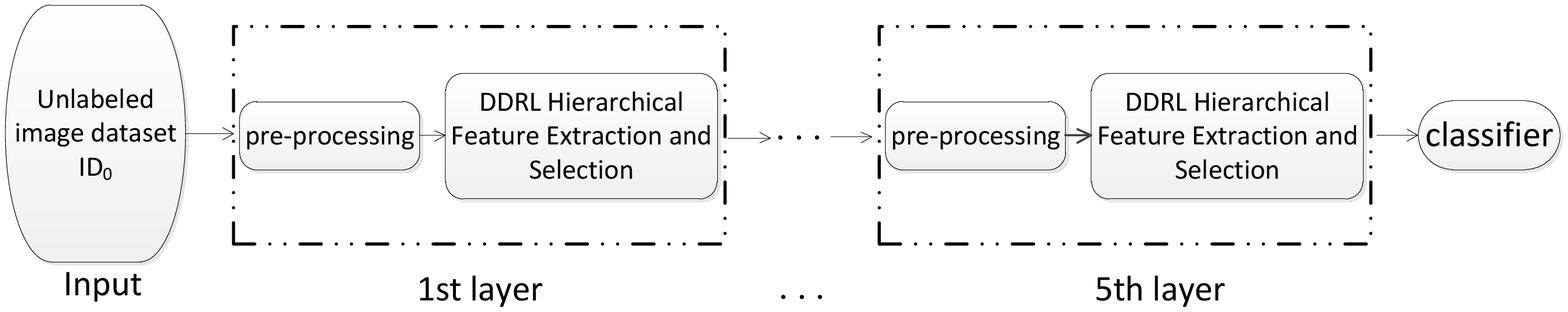}}
\caption{System framework.}
\label{fig1}
\end{figure}

\section{Proposed Approach}
This section presents the detailed
design of our proposed distributed deep representation learning model (DDRL).

\subsection{System Overview}
The structure of DDRL model consists of five layers. As shown in Figure 1, each layer has a similar structure which includes
extracting features and selecting feature maps. The input of the first layer is an unlabeled
image dataset and the output of the last layer, \emph{i.e.} the fifth layer, is the learned
features which are fed to the SVM to test our model. Between the first and the
last layers, each layer extracts features from the output of the previous layer, and then feeds
the extracted features to the next layer. In the process of training model, the training
image set is partitioned into six small datasets $\left\{ID_0, ID_1, \cdots , ID_5 \right\}$
(each of which corresponds to a specified layer) where $\left\{ID_0, ID_1, \cdots , ID_4 \right\}$ are
unlabeled image datasets used to train our DDRL model and $ID_5$ is a
labeled image dataset to train SVM for classification. Here, we divide the whole image dataset to several
small subsets. This step helps  reduce the training time and ensure the richness of the extracted features.

\begin{figure}[H]
\centering
\scalebox{1.2}[1.3]{\includegraphics[width=0.85\linewidth]{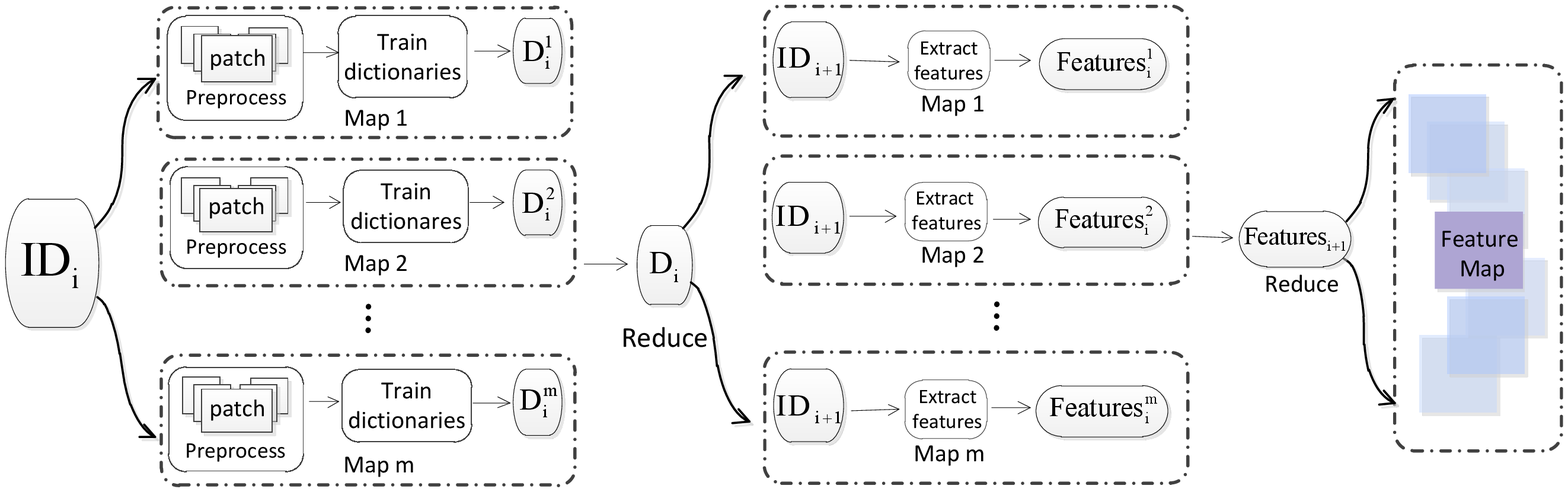}}
\caption{The details of hierarchical feature extraction and selection.}
\label{fig2}
\end{figure}

The specific structure of each layer is depicted in Figure 2, which includes input, pre-processing,
learning dictionary, extracting features and selecting feature maps. Here, we use the first layer as an
example to illustrate our DDRL model. We extract random patches $ID_0[1, \cdots, n]$ from $ID_0$ on the multiple Map
nodes in parallel, followed by a pre-processing stage including contrast normalization and whitening of
these patches. In \cite{Coates9}, it has been proved that normalization can remove the unit limitation of the data, which helps compare and weight the data of different units and whitening helps remove the redundancy between the
pixels. Then, K-means acting as the unsupervised feature learning algorithm runs on each Map node to gain $m$
small dictionaries $\left\{D_{11},D_{21}, \cdots,D_{m1}\right\}$ ($m$ is the total number of map nodes in the cluster),
which will be reduced on the Reduce node to produce the final dictionary $D_1$ of the first layer. Thus, provided the
first layer's feature mapping $\Phi(x;D_1, \varsigma)$, we can extract image features of $ID_1$ and employ a step of
spatial pooling \cite{Boureau14, Boureau15} on these features to obtain more compact representations.
Note that the pooled features are still in rather high dimension and it is hard to select receptive fields from
such huge amount of features. To this end, we utilize a similarity metric to produce feature maps, each of which
contains an equal number of the most similar features, and these feature maps will be input to the second layer and
assigned to the map nodes for parallel processing. Here, on each Map node, K-means will run on the feature maps
to gain the corresponding dictionaries, and subsequent operations are the same as in the first layer.
Finally, in the last layer, we use the dictionary $D_5$ to extract image features of the labeled image dataset $ID_5$ and
then input the pooled features and labels to train the SVM for classification.

\subsection{DDRL Model Formulation}
Suppose that each randomly extracted patch has dimension
$p-by-p$ and has $d$ channels (\emph{e.g.} $d = 3$ for a RGB image), we can then construct a
dataset of $n$ sample patches $X =\left\{X^{(1)}, \cdots ,X^{(n)}\right\}$
, where $X^{(i)} \in R^N$ and $N = p\cdot p\cdot d$.
Given this dataset, the pre-processing step can be done
followed by the unsupervised feature learning process to accomplish the distributed deep representation learning.

\subsubsection{Pre-processing}
Previous state-of-the-art method \cite{Coates9} have validated the key
roles of pre-processing on image patches to improve the subsequent feature learning performance.
In our work, the pre-processing operation involves normalization and whitening, to
provide a cooperative contribution. Since the pre-processing of each image is irrelevant, it can be distributed on the Map node of our DDRL model. Normalization can remove the unit limitation of the data, enabling
comparison and weighting of the data of different units. Whitening helps remove the redundancy between
the pixels. Here, we normalize the patches according to Eq.(1):
\begin{equation}\label{1}
\widetilde{X}^{(i)}=\frac{X^{(i)}-mean(X{^{(i)}})}{\sqrt{var(X^{(i)})+\sigma}},
\end{equation}
where $var(\bullet)$ and $mean(\bullet)$ are the variance and mean of the elements of $X^{(i)}$ and $\sigma$ is a constant added to the variance
before division to suppress noise and to avoid division
by zero. For visual data, this operation corresponds to the local
brightness and contrast normalization.

Since that the adjacent pixel values in the raw input image are highly correlated, we
employ the classical PCA whitening on each $\widetilde{X}^{(i)}$ obtained from the
normalization to make the input less redundant. We have
%\begin{align}
%[\Lambda,U]:=Eig(cov(\widetilde{X})),\label{2} \\
%X^{(i)}_{rot}:=U^{T}\widetilde{X}^{(i)},\label{3}\\
%x_{PCAwhite,i}=\frac{x_{rot,i}}{\sqrt{\lambda_i}},
%\end{align}
\begin{equation}\label{1}
[\Lambda,U]:=Eig(cov(\widetilde{X})),\label{2}
\end{equation}
\begin{equation}\label{1}
X^{(i)}_{rot}:=U^{T}\widetilde{X}^{(i)},\label{3}
\end{equation}
\begin{equation}\label{1}
x_{PCAwhite,i}=\frac{x_{rot,i}}{\sqrt{\lambda_i}},
\end{equation}
We have: where Eq.(2) computes the eigenvalues and eigenvectors of
$\widetilde{X}$ , Eq.(3) uncorrelates the input features, and Eq.(4) obtains the whitened data. Note that
some of the eigenvalues $\lambda_i$ may be numerically close to zero. Thus,
the scaling step where we divide by $\sqrt{\lambda_i}$ would
result in a division by a value close to zero, causing
the data to blow up (take on large values) or otherwise be numerically
unstable. Therefore, we add a small $\epsilon$ to the eigenvalues
before taking their square root, just as shown in Eq.(5).
\begin{equation}\label{5}
x_{PCAwhite,i}=\frac{x_{rot,i}}{\sqrt{\lambda_i+\epsilon}}.
\end{equation}

Furthermore, adding $\epsilon$ here can contribute to smooth the input image,
remove aliasing artifacts caused by the way which pixels laid out in an image, and improve the learned features.

\subsubsection{DDRL Hierarchical Feature Extraction}
Given the pre-processed image data, we input them into the DDRL model to learn hierarchical representation. Specifically, we utilize the K-means algorithm to learn the image statistics and gain a dictionary $D_L$ in the $L$ layer. Then, the pre-processed images $(L=1)$ or the feature maps $(L>1)$ and the duplicated dictionary $D_L$ are distributed on multiple map nodes. On each map node, we define a feature-mapping function $\Phi_L :R_N \to R^{K_L}$ that maps an input vector $x^{(i)}$ to a new feature representation of $K_L$ features. Here, we choose the soft-threshold nonlinearities $\Phi_L(x;D_L, \varsigma) = max\left\{0,D^T_Lx-\varsigma\right\}$, the feasibility of which has been validated in \cite{Coates9}, as our feature extractor, where $\varsigma$ is a tunable constant. Thus, on each map node, we obtain the corresponding feature maps, and K-means is again used to learn dictionaries $D_{L+1}$ from these feature maps. The dictionaries on each map node are then reduced on the reduce node to aggregate a complete one. Similarly, the reduced dictionary is duplicated and distributed on multiple map nodes to respectively extract feature information of the feature maps, just as $L$-layer does. Similar operations are replicated in subsequent layers, and in the last layer, we reduce the learned feature maps into a whole. Section 3.2.3 provides the subsequent operation on these feature maps.

\begin{figure}[H]
\centering
\scalebox{1.1}[1.1]{\includegraphics[width=0.9\linewidth]{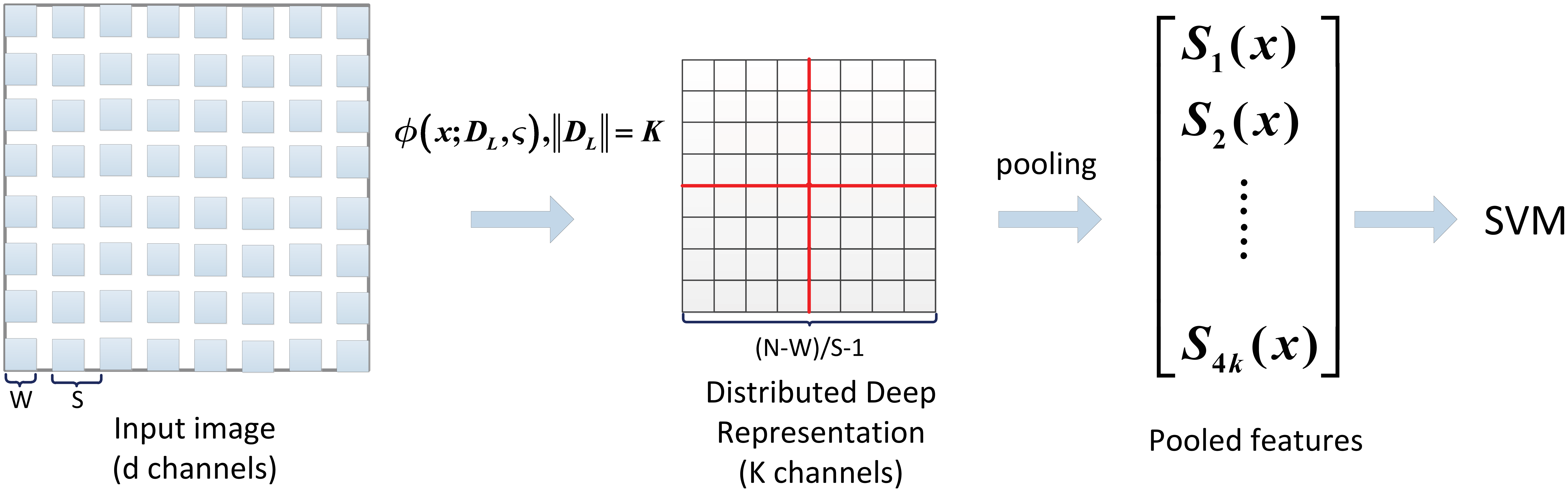}}
\caption{Feature extraction from the input image.}
\label{fig3}
\end{figure}

\subsubsection{DDRL Hierarchical Feature Selection}
Given the feature extractor $\Phi_L$ and $D_L$, we can extract
features $Z_L =\left\{Z_1,\cdots,Z_k\right\}$  of  \(ID_L = \left\{Image_1,\cdots,Image_k\right\}\).
Some previous works [13,14] have theoretically analyzed the importance of spatial pooling to achieve
invariance to image transformation, more compact representations, and better robustness to noise
and clutter. Here, in our work, since that the features learned by K-means are relatively sparse,
we choose average pooling to exploit its advantages.
Figure 3 illustrates the feature extracted from the equally spaced sub-patches
(\emph{i.e.} receptive fields) covering the input image. We first extract $\omega-by-\omega$ receptive
fields separated by S pixels and then map them to $K_L$ dimensional feature vectors via the feature extractor $\Phi_L(x;D_L, \varsigma) = max\left\{0,D^T_Lx-\varsigma\right\}$
to form a new image representation. Then, these vectors are pooled over $2-by-2$ quadrants of the image to form a feature vector for subsequent processing. As
Figure 3 presents, the pooled features will be input to SVM for classification.

Since that the dictionary $D_L$ is designed to be very large
to extract adequate representative features for the accurate
big image data classification, the learned features are commonly in a huge amount
and very high dimensional. Thus, efficiently selecting the receptive fields will be
a rather challenging bottleneck since the single machine may possibly
suffer from a breakdown. On the other hand, even the distributed computing resources
cannot yield a desirable solution because the map nodes just cannot play their full
advantages to process these unorganized features. Therefore, we
wonder what if these unorganized, huge-sized,
and high dimensional features are organized into a whole which can be
easily processed by the map nodes in the cluster? The similarity metric between
features proposed in \cite{Coates16} inspires us to utilize Eq.(6) to produce
feature maps which are composed of equal number of the most similar features.
Given two features $Z_j$ and $Z_k$, the similarity between them is measured as follows:

\begin{equation}\label{6}
d(j,k;Z) \equiv d[Z_j,Z_k]\equiv\frac{\sum_iZ_j^{(i)^2}Z_k^{(i)^2}-1}{\sqrt{\sum_i(Z_j^{(i)^4}-1)\sum_i(Z_k^{(i)^4}-1)}}.
\end{equation}

Here, in our design, the core idea is to find the top $T$ most
correlated features from $Z_L$ as a feature map, and K-means
would then separately take a group of feature maps as input
to obtain the corresponding dictionary $D_L$ on the map nodes
in parallel, which desirably enhances the time efficiency and
avoids the breakdown of the machine.

\section{Experiments}
In this section, we conduct comprehensive experiments to
evaluate the performance of our work on two large-scale
image datasets, \emph{i.e.}, ImageNet \cite{Deng12} and CIFAR-100 \cite{Krizhevsky10}. Here,
we implement a multi-layered network to accomplish the
deep representation learning for subsequent SVM
classification. To provide convincing results, we compare
our work with the method proposed in \cite{Coates9} which similarly
utilized K-means to learn the feature representation on a single
node. To guarantee a fair comparison, we set up the
experimental environment exactly as \cite{Coates9}.

\subsection{Experimental Environment and Datasets}
We built the Hadoop-1.0.4 cluster with four PCs,
each  with 2-core 2.6 GHz CPU and 4 GB memory
space. The total number of the map nodes is 4 and the number of reduce node is 1.

\begin{bfseries}\emph{ImageNet}\end{bfseries} is a dataset of over 15
million labeled high resolution images belonging to roughly 22,000
categories, which aims to provide researchers an easily accessible
image database, and it is organized according to the WordNet hierarchy
in which each node of the hierarchy is depicted by hundreds or thousands
of images. Currently, there is an average of over five hundred images per
node. To validate the performance of our DDRL model, we chose 100 categories, in total 120,000 images from ImageNet datasets, with 80,000 for training and the rest for testing. Since that the images from ImageNet are not of the the same size, we first resized the chosen images to 32-by-32 for the sake of convenience.

\begin{bfseries}\emph{CIFAR-100}\end{bfseries}
consists of 60,000 32-by-32 color images in 100
classes, with 600 images per class. There are 500
training images and 100 test images per class.

%Sample images from CIFAR-100 and ImageNet
%are presented in Fig.4 and Fig.5, respectively.

\begin{figure}[H]
\centering
\scalebox{1.2}[1.2]{\includegraphics[width=0.8\linewidth]{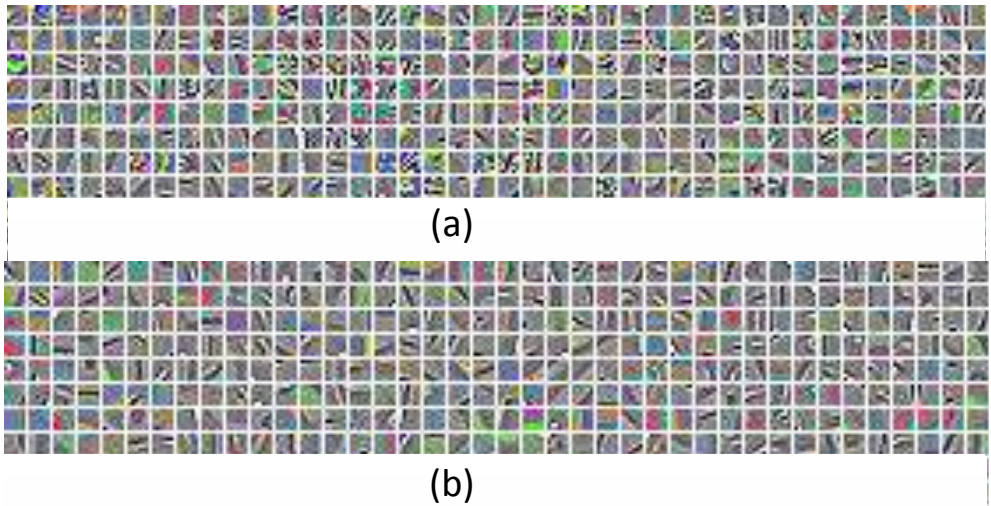}}
\caption{CIFAR-100 feature extraction. (a): a random selection of 320 filters chosen
from the 6-by-6-by-1600 image representations learned by the first layer of DDRL model.
(b): a random selection of 320 filters (out of 1600) of size 6-by-6 learned by \cite{Coates9}.
}
\label{fig56}
\end{figure}
\subsection{Comparison of Dictionary}
Before looking at the classification results, we first
inspect the dictionary learned by our DDRL model and the
dictionary learned by \cite{Coates9} on CIFAR-100. The receptive
field size is $6\times 6$ and the stride between two receptive
fields is $1$ pixel.  As presented in Figure 4, (a)
provides the randomly selected filters (out of $1,600$)
learned by the first layer of DDRL model, and (b) gives
a random selection of filters from the completed dictionary
composed of $1,600$ filters learned by \cite{Coates9}. Visually,
little difference can be observed between (a) and (b). Both of them
present diversified features to contribute to the subsequent
feature extraction and SVM classification. It is worth to
mention that the dictionary presented in (a) is gained by
reducing the four dictionaries obtained on the four map nodes,
and (b) is gained using a single machine. Thus, considering both
the computing resources and the final obtained similar dictionary,
we can demonstrate that the feature learning model proposed in our
work is superior to that presented in \cite{Coates9}. If we need to learn a much
bigger dictionary for better classification performance, the approach proposed in \cite{Coates9} will
impose a serious computation constraint on the single machine while
our distributed deep model (DDRL) is competent to tackle this
challenge with the joint efforts of the distributed
computing resources.

\begin{figure}[H]
%\centering  \linewidth
\centering
%\fbox{
\scalebox{1.1}[1.1]{\includegraphics[width=0.5\linewidth]{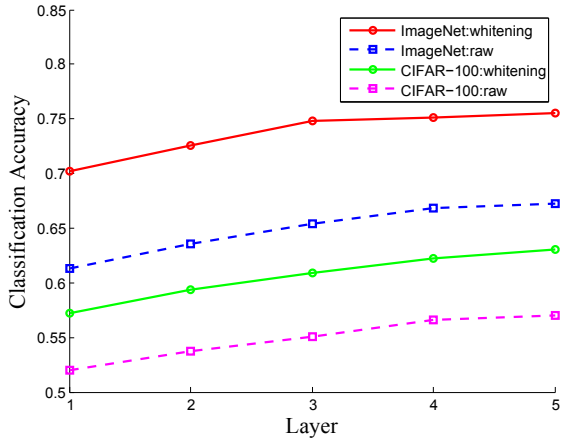}}
\caption{Effect of whitening.}
\label{whitening}
\end{figure}
\subsection{Effect of Whitening}
Whitening is a part of pre-processing, which can help remove the redundancy of the pixels. In this section, we conduct experiment to validate the effect of whitening on our model.

Figure 5 shows the performance of our model with and without whitening. Our DDRL model has 5 layers, with 1,600/2,000/2,400/2,800/3,200 centroids from the first layer to the last. In all these layers, the size of the receptive filed is $6\times 6$, and the stride is set as $1$ pixel. From the experimental results, both on ImageNet and CIFAR-100 dataset, we observe that the performance of our model gets improved when the layer number increases, and this increase takes place no matter whether or not the whitening operation is included. The reason for this increase will be discussed in the next subsection. In addition, Figure 5 shows that when the model depth is the same, the whitening operation can help DDRL model achieve higher classification accuracy, both on ImageNet and CIFAR-100 dataset. Thus, we can conclude that whitening is a crucial pre-processing to optimize the proposed model.

\begin{figure}[H]
\centering
\scalebox{1.1}[1.1]{\includegraphics[width=0.5\linewidth]{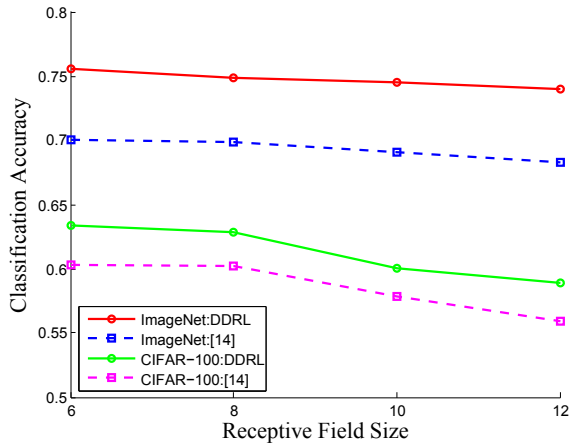}}
\caption{Effect of receptive field size.}
\label{rfSize}
\end{figure}

\subsection{Effect of Receptive Field Size and Stride}

In this section, we conduct experiments to compare the effect
of receptive field size and stride on DDRL model and \cite{Coates9}, both on ImageNet and CIFAR-100 dataset.

Figure 6 illustrates the effect of receptive field size
between DDRL model and \cite{Coates9} on ImageNet and CIFAR-100 dataset.
The result of \cite{Coates9} is
gained with stride=1 pixel and 1,600 centroids.
The results of DDRL model is obtained with 5 layers
(centroids number per layer is 1,600, 2,000, 2,400, 2,800, 3,200)
and 1 pixel stride. In this experiment, we set the receptive
field size as $6\times6$, $8\times8$, $10\times10$, and $12\times12$.
As the lines present, both DDRL model and \cite{Coates9} show decreasing performance
when the receptive field size increases, while DDRL model
still achieves higher accuracies than \cite{Coates9} in all cases. From this
perspective, smaller receptive field will lead to better
performance, which, however, will result in higher
computation expense. In this sense, both on ImageNet and CIFAR-100 dataset, our model can release
such constraint with the distributed storage and computing,
while with the approach proposed in \cite{Coates9}, it is hard to deal with this overhead.
\begin{figure}[H]
%\centering  \linewidth
\centering
%\fbox{
%\includegraphics[width=0.85\textwidth]{stride.pdf}
\scalebox{1.1}[1.1]{\includegraphics[width=0.5\linewidth]{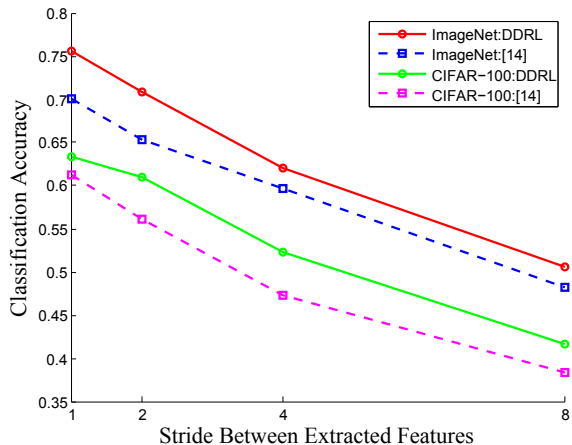}}
\caption{Effect of stride.}
\label{Stride}
\end{figure}

Figure 7 presents the effect of different strides proposed model in \cite{Coates9} and DDRL model on ImageNet and CIFAR-100 dataset. The model in \cite{Coates9} sets the receptive field size as $6\times6$ and centroids number
as 1,600. Our DDRL model is the same as described before, consisting of 5 layers
with 1,600/2,000/2,400/2,800/3,200 centroids at different layers
and the receptive field size is fixed at $6\times6$. Similar to Figure 6,
both the model in \cite{Coates9} and DDRL model get decreasing performance when the stride
increases, and DDRL model keeps superiority over \cite{Coates9}
at all stride values. Similarly, smaller stride makes great
contribution to the classification performance while introducing
extra computation cost. Again, on ImageNet and CIFAR-100 dataset,
our DDRL model can overcome the
computational constraint with the distributed computing resources
while it is difficult for a single machine to overcome such a problem.
\begin{table}[H]
 \caption{Comparison of the classification performance on ImageNet dataset.}
  \centering
  \begin{tabular}{|p{1.3cm}|p{1.2cm}|p{1.2cm}|p{1.2cm}|p{1.2cm}|p{1.2cm}|p{1.2cm}|}
    \hline
      layer &  1 &  2 &  3 &  4 &  5\\ \hline
       \textbf{DDRL}   &  70.19\%   &  72.58\% &  74.86\% &  75.14\% &  75.53\% \\\hline
       \cite{Coates9} &  70.01\%   &  N/A  &  N/A  &  N/A  &  N/A  \\\hline
  \end{tabular}
  \label{table1}
\end{table}

\begin{table}[H]
 \caption{Comparison of the classification performance on CIFAR-100 dataset.}
  \centering
  \begin{tabular}{|p{1.5cm}|p{1.2cm}|p{1.2cm}|p{1.2cm}|p{1.4cm}|}
    \hline
      Method & \cite{Coates9} & \cite{Goodfellow30} & \cite{Nitish31}& \textbf{DDRL}\\ \hline
      Accuracy& 61.28\%  &61.43\%  &63.15\%  & \textbf{62.83\%} \\\hline
  \end{tabular}
  \label{table1}
\end{table}

\subsection{Classification Performance }
In this section, we validate the classification performance of our DDRL model on ImageNet and CIFAR-100. In both \cite{Coates9} and DDRL model, the receptive field size is $6 \times 6$, and the stride is $1$. Table 1 presents the results we gained on models with different number of layers. Although the inherent map and reduce phase of MapReduce may inevitably bring some compromises, the superiority of DDRL model become obvious when the layer number grows. A two-layer setup (with 1,600 centroids in the first dictionary and 2,000 centroids in the second) lead 2.39\% improvements on ImageNet compared with the single layer. The results gained on a three/four/five-layer model continue to achieve an increase to different extent. The five-layer model gained subtle increase (only 0.39\%) compared with the four-layer one, which indicates that when the model reaches an enough depth, the classification performance will gradually stop improving. Considering the consumption of computation and storage resources, the five-layer depth of DDRL model is deep enough in general.

Although the main body of work is conducted on the ImageNet dataset, we also investigate how the model performs on the CIFAR-100 dataset. As shown in Table 2, our DDRL model achieves 62.83\% accuracy. We can observe that the result of DDRL model outperforms \cite{Goodfellow30} by 1.4\%, and \cite{Coates9} by 1.55\%. Compared to \cite{Nitish31}, DDRL model gets 0.32\% lower accuracy, which was mainly resulted from the required image amount to train the classification model. Thus, some relatively small
datasets (\emph{e.g.} NORB and CIFAR-10, \emph{etc..}) were not used to validate the performance of our DDRL model. Considering both the classification results and computing consumption, such subtle discrepancy is acceptable and reasonable. Thus, when the amount of training images is large enough, the superiority of our DDRL model may become more obvious.

\section{Conclusion}
In this work, we have successfully implemented the distributed deep representation
learning model (DDRL) focusing on the task of big image data classification. Different from previous methods perusing algorithm optimization, our design focuses on employing some elegant designs to enhance the classification accuracy while
 retaining the simplicity of the feature learning algorithm and the robustness
in the implementation platform. Since that desirable accuracy of big image data
classification imposes a high requirement for the amount and
richness of features, we proposed a platform with excellent fault tolerance
to avoid the breakdown of the single
machine. Experimental results demonstrated the encouraging
performance of our design and we expect to pursue tackling further challenges in the big image data in the future work.

\section*{Acknowledgments}
This work was supported in part by the National Natural Science Foundation of China under Grant 61370149, in part by the Fundamental Research Funds for the Central Universities(ZYGX2013J083), and in part by the Scientific Research Foundation for the Returned Overseas Chinese Scholars, State Education Ministry.

\end{document}